\documentclass[10pt,twocolumn,letterpaper]{article}

\usepackage{iccv}
\usepackage{times}
\usepackage{epsfig}
\usepackage{graphicx}
\usepackage{amsmath}
\usepackage{amssymb}
\usepackage{multirow}

\usepackage[pagebackref=true,breaklinks=true,letterpaper=true,colorlinks,bookmarks=false]{hyperref}

\iccvfinalcopy 

\def\iccvPaperID{8169} 

\ificcvfinal\fi

\begin{document}

\title{Noise Modulation: Let Your Model Interpret Itself}

\author{Haoyang Li\ \ \ \ Xinggang Wang\\
Huazhong University of Science and Technology\\
Wuhan, Hubei, China\\
{\tt\small \{lihaoyang,xgwang\}@hust.edu.cn}
}

\maketitle
\ificcvfinal\thispagestyle{empty}\fi

\begin{abstract}
   Given the great success of Deep Neural Networks(DNNs) and the black-box nature of it, the interpretability of these models becomes an important issue. 
   The majority of previous research works on the post-hoc interpretation of a trained model. But recently, adversarial training shows that it is possible for a model to have an interpretable input-gradient through training.
   However, adversarial training lacks efficiency for interpretability.
   To resolve this problem, we construct an approximation of the adversarial perturbations and discover a connection between adversarial training and amplitude modulation. Based on a digital analogy, 
   we propose noise modulation as an efficient and model-agnostic alternative to train a model 
   that interprets itself with input-gradients. 
   Experiment results show that noise modulation can effectively increase the interpretability of 
   input-gradients model-agnosticly.
\end{abstract}

\section{Introduction}

\begin{figure}[t]
   \begin{center}
      \includegraphics[width=0.8\linewidth]{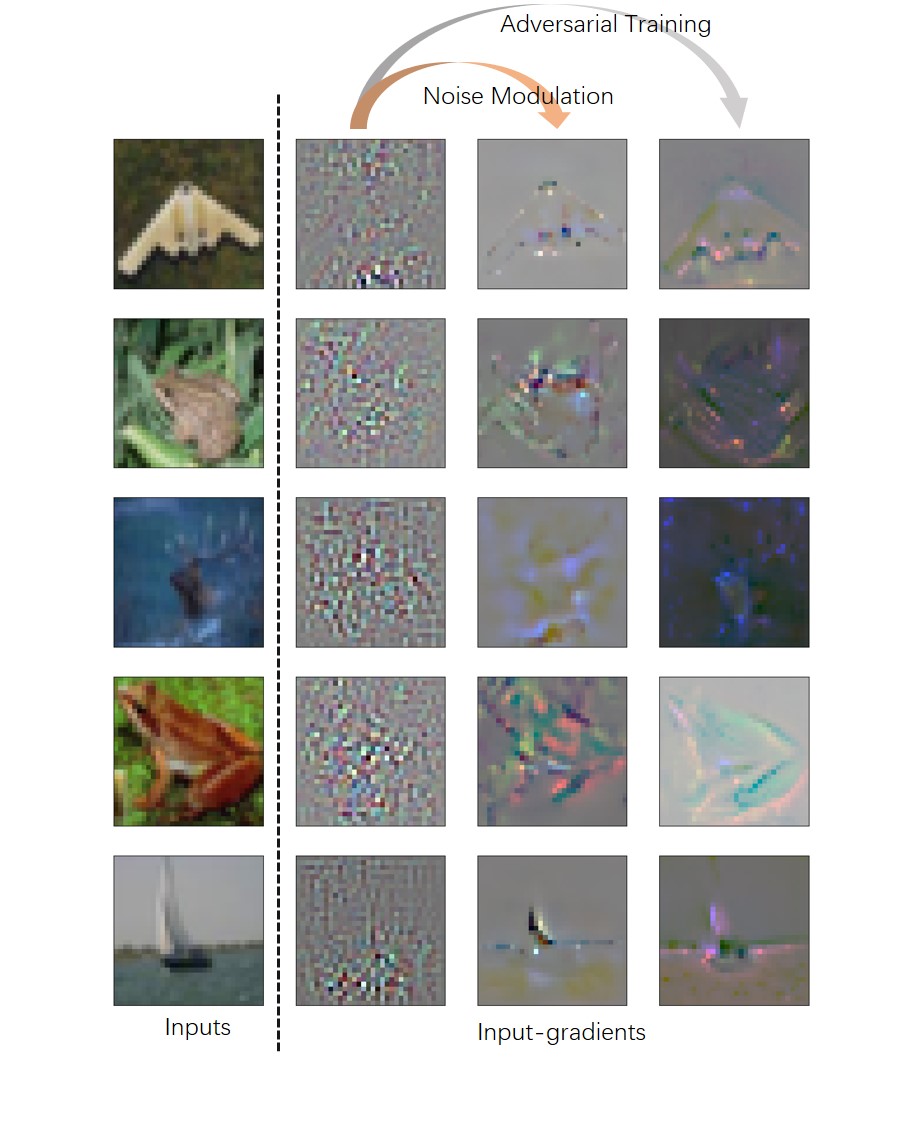}
   \end{center}
      \caption{Visualization of the input-gradients, \ie the loss gradient with respect to input pixels.
      The gradients are scaled to $[0,1]$ for visualization.
      The first column are original inputs.
      The second, third and fourth columns are input-gradients of 
      the same model acquired with standard training, \emph{noise modulation} and adversarial training respectively.
      We propose an efficient and model-agnostic alternative to recover the interpretable input-gradients brought by adversarial training.
      }
   \label{fig:demo}
\end{figure}

Deep Neural Networks (DNNs) have demonstrated huge potential in solving multiple visual recognition problems, \eg, image classification~\cite{Alex2012imagenet}, object detection~\cite{Zhengxia2019objectdetection} and  segmentation~\cite{Georgios2020segmentation}. 
But the back-box nature of them hinders their further application in scenarios where the decisions can result in danger, such as medical prognosis, autonomous driving and verification~\cite{Avanti2016gradinput}.
Given that these successful models are now being deployed into daily products, 
it's important to explain their predictions to ensure reliability, robustness and fairness~\cite{Ancona2019gradient}. 

There are three major approaches to resolve this lack of interpretability~\cite{Ancona2019gradient}, 
the first is to use an inherently interpretable model, \eg linear model, which results a limited predictive power; 
the second is to design a model that generates predictions and explanations together, 
which is challenging due to the lack of ground-truth explanations; the third is 
to give post-hoc interpretations for a trained model, which is the prevailing approach~\cite{Ancona2019gradient}.

The study of interpretability also leads to the discovery of adversarial example~\cite{Christian2014intriguing}. 
Based on this discovery, adversarial attacks, \ie generating adversarial examples to cheat the targeted model, have evolved into a striking threat against systems equipped with these models~\cite{Cihang2017adversarial}. 
To defend from these attacks, many methods have been proposed and breached in the past few years \cite{Anish2018obfuscated}~\cite{Warren2017breachensemble}~\cite{Nicholas2017bypassing}. 
It is adversarial training~\cite{Ian2015explaining}~\cite{Madry2018towards}, \ie augmenting training set with adversarial examples,  
that survives through sanity checks and adaptive attacks~\cite{Anish2018obfuscated} and becomes the leading defense method till today. 

When \cite{Madry2019robustnessodds} checks the models acquired through adversarial training, it is surprising that these models now have a clear and interpretable input-gradient, \ie the loss gradient with respect to inputs, as further confirmed by ~\cite{Tianyuan2019intrepreting}.
It occurs to us that adversarial training actually gives a regular model 
the ability to generate interpretable input-gradients, in other words, it is possible 
to train a regular model that interprets itself with input-gradients~\cite{Christian2019connection}.

But the sideeffects of adversarial training is overwhelming, 
as reported in literature, it requires more data~\cite{Ludwig2018moredata}, larger model ~\cite{Cihang2020intriguing} 
and much more computations to achieve an accuracy comparable but inferior to standardly trained models~\cite{Madry2019robustnessodds}. 
It lacks of efficiency in terms of interpretability.
This leads us to the following question we study in this paper:
\emph{Is there an efficient alternative to train a model that interprets itself with input-gradients?}

To address this question, we start by looking for an efficient approximation of adversarial perturbations used for adversarial training.
It surprisingly links to the technique of \emph{amplitude modulation}.
In communication, \emph{amplitude modulation} refers to having the amplitude of the carrier wave vary in proportion to that of the message signal.
Based on a digital analogy, we propose \emph{noise modulation} as an efficient and model-agnostic alternative to train a model with interpretable input-gradients.

Our contributions are summarized as follows:
\begin{itemize}
   \item We discover a connection between adversarial training and amplitude modulation through an approximation of adversarial perturbations. It offers a new perspective to understand the effects of adversarial training on input-gradients.
   \item We propose an efficient and model-agnostic alternative, namely \emph{noise modulation}, to train a model that interprets itself with clear and human-aligned input-gradients.
\end{itemize}


\section{Motivation \& Approach}\label{sec:motivation}

\subsection{How Adversarial Training Works}

Adversarial training was first proposed in \cite{Ian2015explaining}, 
but the most popular and effective formulation is proposed by ~\cite{Madry2018towards}.
The core idea of adversarial training is to add adversarial examples into training set, \cite{Madry2018towards} even substitute the whole training set with adversarial examples.

Given a dataset $D=\{(x,y)|x\in \mathbb{R}^N, y\in \mathbb{N}\}$ 
containing pairs of input $x$ and label $y$, a model $f$ parametrized with $\theta$ and a loss function $L$, adversarial training refers to solving 
the following minimax optimization problem. Following the formulation of \cite{Madry2018towards},
\begin{equation}\label{equ:pgd_at}
\min_{\theta}
\mathbb{E}_{(x,y)\sim\mathcal{D}}\left[\max_{\delta\in B(\varepsilon)}L(f(\theta;x+\delta),y)\right].
\end{equation}

At each step, the inner maximization first searches for an optimal perturbation $\delta^*$ in a given attack space $B(\varepsilon)$, \eg a $l_{\infty}$-norm ball 
$\{\delta|\delta\in\mathbb{R}^N,\varepsilon\in\mathbb{R},||\delta||_{\infty}\le\varepsilon\}$, such that the loss function is maximized; 
the outer minimization then optimizes the parameters $\theta$ to minimize 
the loss function just like standard training, but on perturbed examples $x+\delta^*$. 

The inner optimization is commonly solved using Projected Gradient Descent (PGD) attack and its variants. It is an iterative method that at each epoch $t\in[T]$ ($T$ epochs in total), the perturbation is moved towards the direction of input-gradient for a certain step and projected back to the feasible space, \ie
\begin{equation}\label{equ:pgd_attack}
\begin{aligned}
\delta^{t+1}=P_{B(\varepsilon)}(\delta^t+\alpha\text{sign}(\nabla_{x+\delta^t}L(f(\theta;x+\delta^t),y)))\\
\delta^0\sim \mathcal{U}(-\varepsilon,-\varepsilon),
\end{aligned}
\end{equation}
where $P_{B(\varepsilon)}$ denotes projecting the perturbations back to the attack space, $\alpha\in\mathbb{R}$ denotes the step size at each iteration.
After $T$ epochs, the adversarial example $x^*=x+\delta^*$ perturbed with the optimal perturbations $\delta^*$,
is further fed for training. It only changes the training strategy, at inference, 
the model still gives its predictions on clean inputs.

The  heavy computations of adversarial training
mainly stems from the computation of input-gradients at each iteration of PGD.
For adversarial training with $T$ epochs of PGD,
there are $T$ more back-propagation iterations computed than a standard training.
Other sideeffects of adversarial training, \eg requiring larger model and more data 
comes from the adversarial nature of these perturbations.
These perturbations are designed to move the original examples towards 
the decision boundary, making the resulted adversarial examples much more difficult to fit for model~\cite{Madry2019robustnessodds}.
The increased difficulty requires a larger model to fit given the same amount of data~\cite{Cihang2020intriguing}, this further increases the computational overheads as
the computation of back-propagation also grows with the scale of model. 

\subsection{Approximation of Adversarial Perturbations}

\begin{figure*}[t]
   \begin{center}
      \includegraphics[width=0.8\linewidth]{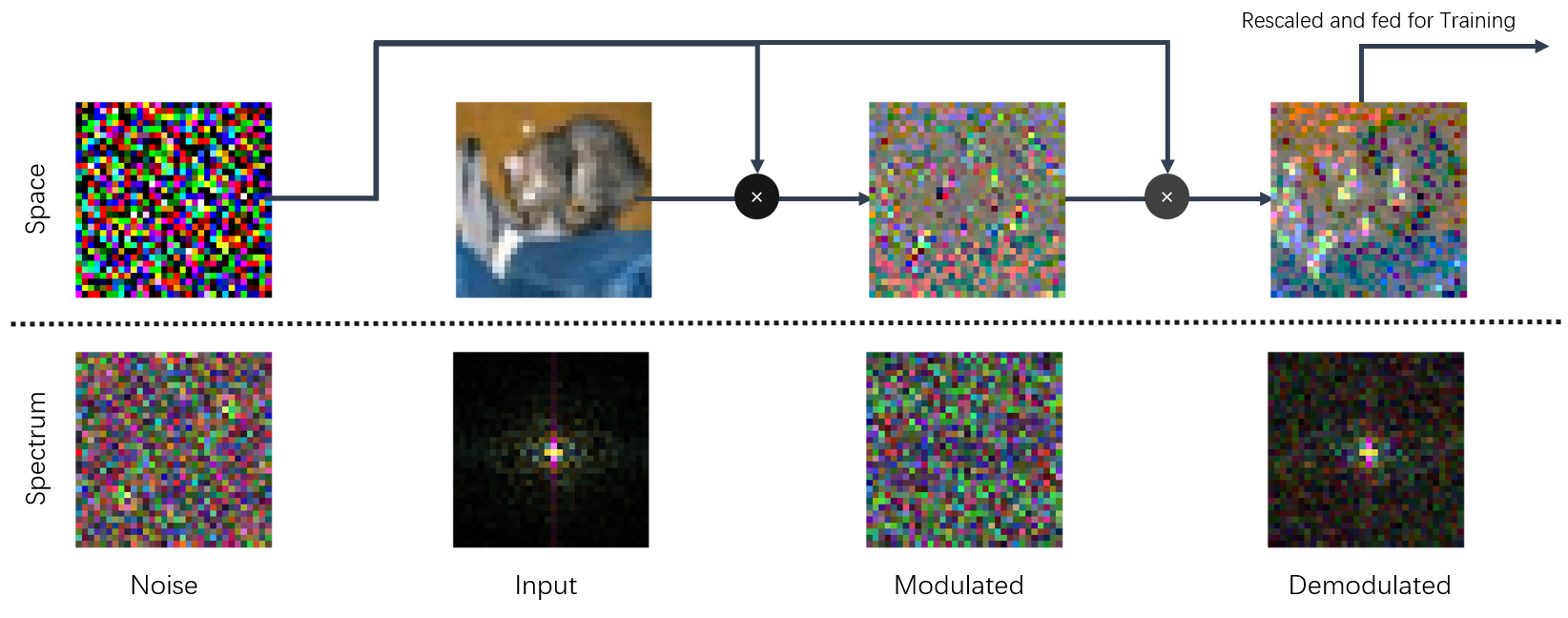}
   \end{center}
      \caption{The process of noise modulation.  For each input, a noise is sampled first and then multiplied with input twice.  The demodulated input is rescaled to keep the constant component stasis and then fed for training.      We pictured the change of inputs in space and spectrum   (the spectrum is scaled to $[0,1]$ for better visualization).   The discriminative feature is hidden on amplitude after modulating, but then recovered by demodulating.}
   \label{fig:nm_structure}
\end{figure*}

In this section, we will draw an approximation of the adversarial perturbations 
in an efficient way. The first step, as initially proposed by \cite{Ian2015explaining}, is to substitute the iterative multi-step PGD attack with its one step approximation, \ie
\begin{equation}\label{equ:fgsm_attack}
\begin{aligned}
   \delta^*\sim P_{B(\varepsilon)}(\delta^0+\alpha\text{sign}(\nabla_{x+\delta^0}L(f(\theta;x+\delta^0),y))).
\end{aligned}
\end{equation}

The perturbation now becomes a projected linear combination of a uniform noise and the sign of input-gradient. Our second step is to find a way to approximate the input-gradient. It has been observed that adversarial training makes 
the input-gradients similar to the corresponding inputs~\cite{Alvin2020Jacobian}.
This can be verified by visualizing the input-gradients of adversarially trained model, as shown in the fourth column in Figure~\ref{fig:demo}.

This similarity drives us to hypothesize that the sign of input-gradient 
can be approximated roughly by a multiplication of the corresponding input and some noise $\delta^\prime\in\mathbb{R}^N$
\footnote{Apparently, the sign of input-gradient is not the same as corresponding input.}, \ie $\text{sign}(\nabla_x L)\sim \delta^\prime\cdot x$.
Take it back to Equation~\ref{equ:fgsm_attack}, the perturbation now becomes
\begin{equation}\label{equ:grad_approx}
\begin{aligned}
   \delta^* \sim P_{B(\varepsilon)}(\delta+\alpha\cdot\delta^\prime\cdot x).
\end{aligned}
\end{equation}

This approximation of Equation~\ref{equ:grad_approx} actually estimates the adversarial perturbation with a modified input. It is approximately the case after adversarial training, as reported in \cite{Zeyuan2020feature}. If we take this approximation back into adversarial training, the example $x^*$ fed for training becomes
\begin{equation}\label{equ:partial_approx}
\begin{aligned}
   x^*&= x + \delta^* \\
      &\sim P_{x+B(\varepsilon)}(x\cdot(1 + \alpha\cdot\delta^\prime)+\delta).
\end{aligned}
\end{equation}

For our purpose, the perturbations are not intended to be adversarial,
therefore, we further remove the components that are irrelated to input.
The final approximation will be
\begin{equation}\label{equ:pgd_approx}
   x^*\sim x\cdot(1 + \alpha\cdot\delta^\prime).
\end{equation}

This formulation reminds us of the technique of \emph{amplitude modulation}.
In communication, \emph{amplitude modulation} refers to 
having the amplitude of a carrier signal, generally of high-frequency, vary 
in proportion to that of the message signal, \ie the signal that contains the information we wish to transmit,
before transmission.
At the receiver, the modulated signal is first demodulated with carrier of the same frequency, and then filtered through a low-passing filter to recover the original signal. From this perspective, adversarial training works as a dynamic modulator 
that tunes the model to filter out the informative components of frequencies by itself.

It motivates us to rethink this formulation from the perspective of modulation, 
and leads us to \emph{noise modulation} we will propose in the next section.

\subsection{Noise Modulation}

For a digital signal like an image, we can also modulate it over a digital carrier of the same size and then demodulate the modulated signal using the same carrier.
If we feed this demodulated signal for training, the model will have to learn to filter out informative components of frequencies related to its purpose. Based on this digital analogy of \emph{amplitude modulation}, we propose \emph{modulational training} as formulated below.

\newtheorem{definition}{Definition}
\begin{definition}\label{def:modulational_training}
\emph{\textbf{Modulational Training}}
Given a dataset $D=\{(x,y)|x\in \mathbb{R}^{N},y\in \mathbb{N}\}$,
a model $f$ prametrized with $\theta$,
a loss function measuring the distance between predictions $f(x)$ and true label $y$,
a carrier $c\in\mathbb{R}^N,c\neq 0$, modulational training refers to solving the following optimization problem
\begin{equation}\label{equ:mt_def}
\begin{aligned}
\min_{\theta}\mathbb{E}_{(x,y)\sim\mathcal{D}}\left[L(f(\theta;x\cdot\frac{N\cdot c^2}{C_0}),y)\right]\\
C_0=\sum_{n=0}^{N-1}c_n^2,
\end{aligned}
\end{equation}
when $c=\mathbf{1}$, it becomes standard training.
\end{definition}

It seems quite different from Equation~\ref{equ:pgd_approx}, but 
we will show that they are the same in essence.
For the second power of a non-zero carrier $c^2=\{c_n^2\},c_n\in\mathbb{R}$, 
the $k$-th component of its Discrete Fourier Transform (DFT) $C=\{C_k\}, C_k\in\mathbb{C}$ is
\begin{equation}\label{equ:c_dft}
C_k=\sum_{n=0}^{N-1}c_n^2\text{e}^{-j\frac{2\pi k n}{N}}.
\end{equation}

The constant component $C_0=\sum_{n=0}^{N-1}c_n^2$ is positive,
as long as $c \neq 0$. 
Thus the second power of this non-zero carrier can be rewritten as 
a combination of a constant $\lambda=\frac{C_0}{N}$ and 
some noise $\delta^{\prime\prime}\in\mathbb{R}^N$,i.e.
\begin{equation}\label{equ:c_analysis}
c^2=\lambda+\delta^{\prime\prime}\implies x\cdot\frac{N\cdot c^2}{C_0} =x\cdot(1+\delta^{\prime\prime}\cdot\frac{N}{C_0}).
\end{equation}

This formulation links back to our approximation of adversarial training (Equation~\ref{equ:pgd_approx}).
It indicates that it is possible to recover the original signal by filtering out the extra noises. Since the filtering of a certain frequency is a linear operation,
in fact, a convolution, it is learnable by DNNs 
and even better for Convolutional Neural Networks(CNNs).

We still do not know which carrier can completely recover the interpretable input-gradients brought by adversarial training. 
But as we strives for efficiency, we propose to use noise as carrier and 
name this special case of \emph{modulational training} as \emph{noise modulation}.

\begin{definition}\label{def:noise_modulation}
\emph{\textbf{Noise Modulation}}
Given a a noise $\delta\in\mathbb{R}^N$, \eg standard Gaussian noise $\delta\sim\mathcal{N}(0,\mathbf{I}_{N})$,
noise modulation refers to solving the following optimization problem
\footnote{the rest notations are the same with those in Definition~\ref{def:modulational_training}.}.
\begin{equation}\label{equ:nm_def}
\begin{aligned}
   \min_{\theta}\mathbb{E}_{(x,y)\sim\mathcal{D}}\left[L(f(\theta;x\cdot\frac{N\cdot c^2}{C_0}),y)\right]\\
   c=\beta + (1-\beta)\cdot \delta, C_0=\sum_{n=0}^{N-1}c_n^2,
\end{aligned}
\end{equation}
where $\beta\in[0,1]$ is a hyper-parameter that controls the ratio of constant component in carrier $c$.
When $\beta=1$, it becomes standard training.
\end{definition}

The process of \emph{noise modulation} is illustrated in Figure~\ref{fig:nm_structure}.
For each input, a noise is sampled and then multiplied with the input twice.
The demodulated input is rescaled to keep the constant components stasis and then fed for further training.
The modulated input hides the discriminative features on the amplitude and 
these features are then brought back by demodulation.
At inference, there is no processing for inputs, just like adversarial training.

As shown in Definition~\ref{def:noise_modulation}, \emph{noise modulation} requires 
no extra back-propagation iterations than standard training. 
Since the noise is sampled independently from model, 
it is model-agnostic by design and can be further accelerated by prepossessing the dataset before training.
Its computation grows with the dimension of data rather than the scale of model.
\emph{Theoretically, it is much more efficient than adversarial training.}

\section{Experiments}\label{sec:experiments}

\begin{figure*}[t]
   \begin{center}
      \includegraphics[width=0.9\linewidth]{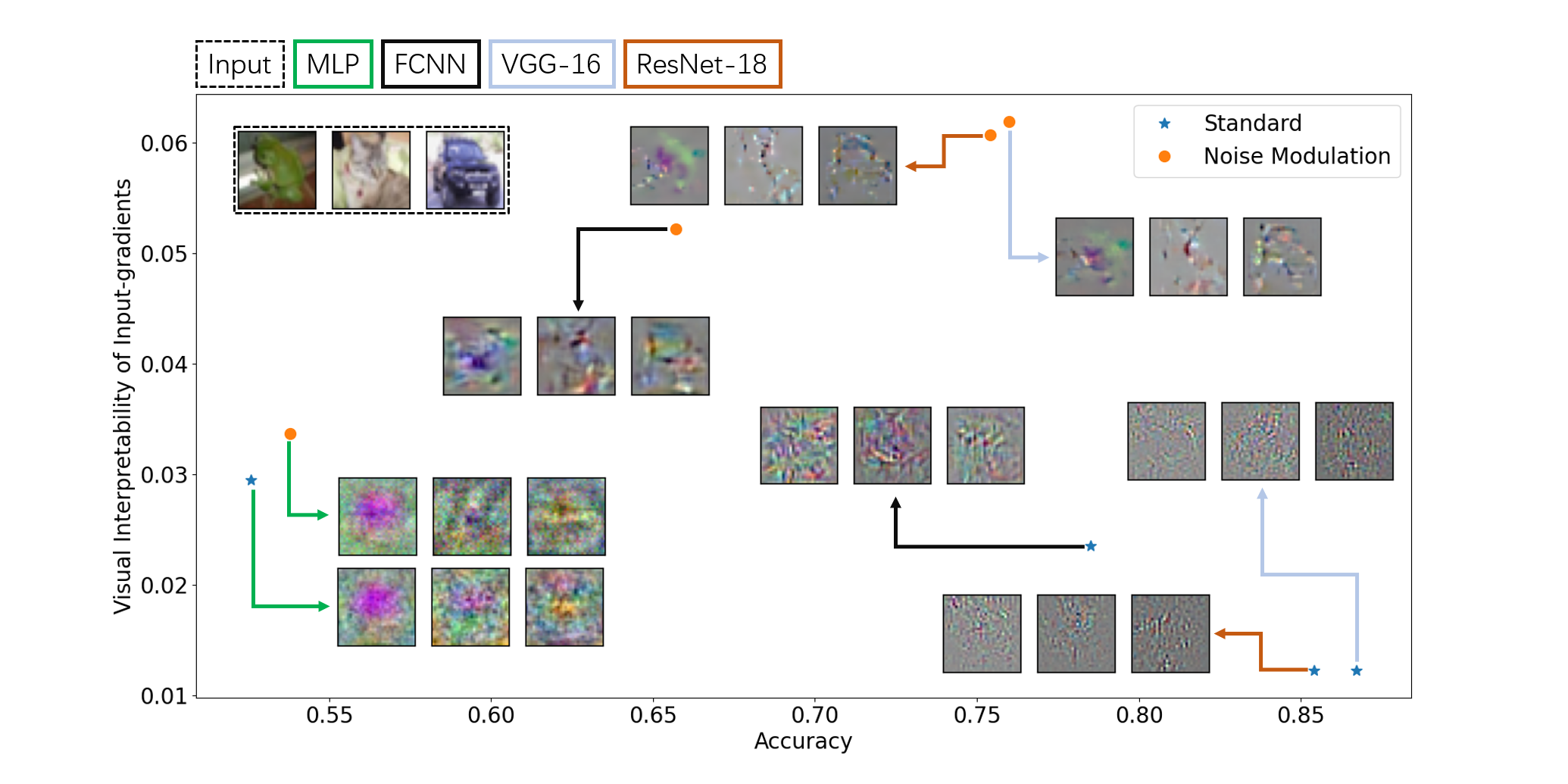}
   \end{center}
   \begin{center}
         \includegraphics[width=0.9\linewidth]{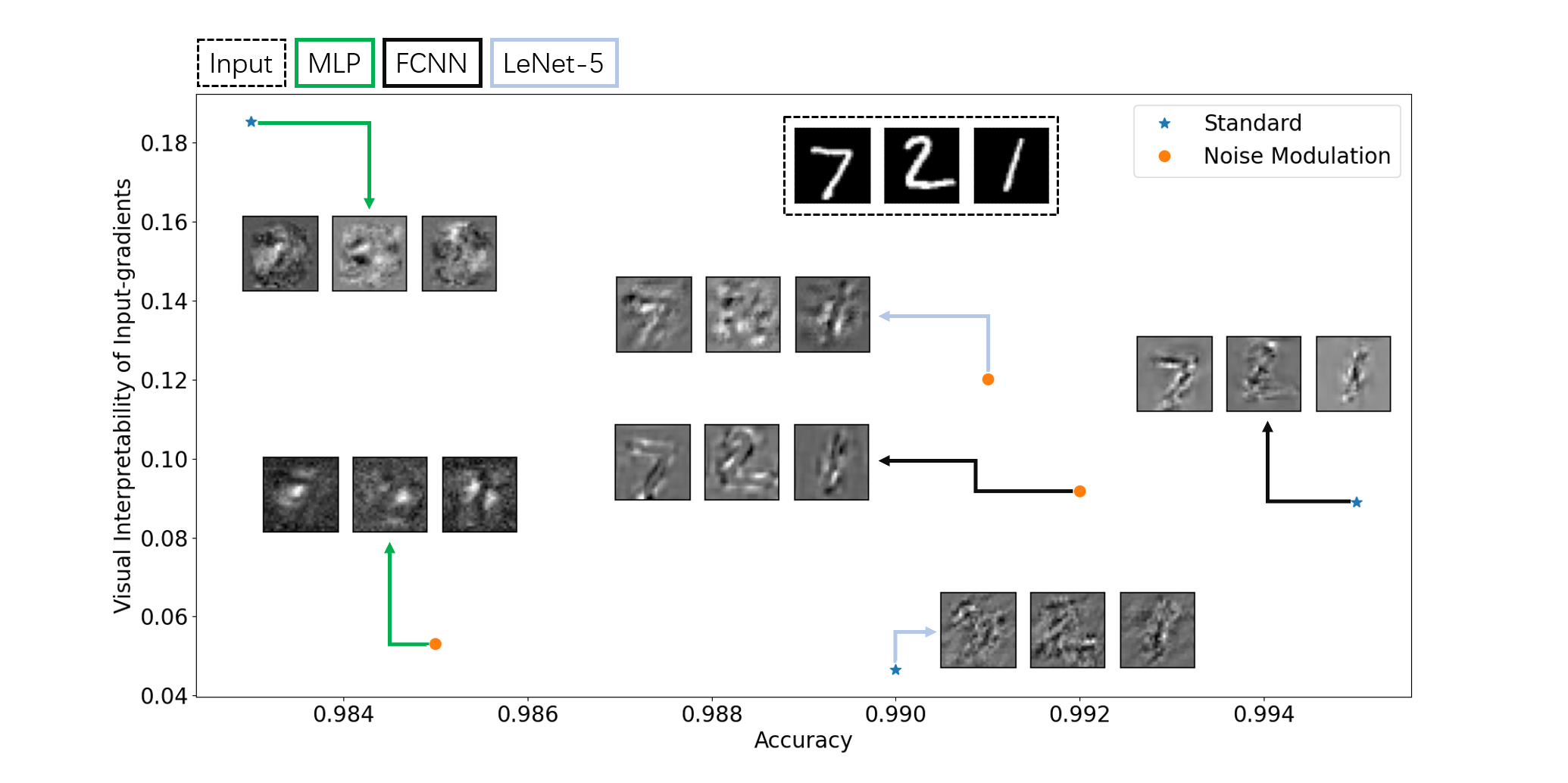}
      \end{center}
   \caption{Experiment results on CIFAR-10 (up) and MNIST(bottom). 
            Each color marks a model. The dotted frame marks for the inputs. 
            Noie modulation improves the interpretability of input-gradients significantly.}
   \label{fig:vis_comparison}
\end{figure*}

In this section, we will first define a metric to measure the visual interpretability of input-gradients such that we can validate the effectiveness of \emph{noise modulation} on input-gradients both qualitatively and quantitatively.
We will then show the trade-off between interpretability and accuracy, and the 
influences of different choices of noises used in \emph{noise modulation}.

All of the following experiments are conducted using PyTorch~\cite{paszke2019pytorch} with a random seed fixed at $0$.
The hyper-parameters are set as the default unless specially mentioned.

\subsection{Evaluating Input-gradients}

In terms of interpretable input-gradients, we mean that 
the input-gradient of model serves as an attribution map telling us 
how each dimension of an input influences the model's decision.
As proved by the existence of adversarial examples ~\cite{Christian2014intriguing},
moving the input towards the direction of its corresponding input-gradients increases the loss function.
It indicates that input-gradients contain information crucial to the decision of model 
while not necessarily making sense for human.

Qualitatively, the input-gradient is more interpretable if it is more visually aligned with input~\cite{Christian2019connection}.
Given the fact that numerically the input-gradient is much more smaller than input, 
we will use the absolute cosine similarity over their signs,
to define a Visual Interpretability of Input-gradient(VII) as follows.

\begin{definition}\emph{\textbf{Visual Interpretability of Input-gradient (VII)}}
Given a differentiable model $f$ and a loss function $L$,\eg cross-entropy , measuring the distance between its prediction and ground truth, 
its visual interpretability of input-gradients over 
a test dataset $\mathcal{D}=\{(x,y)|x\in \mathbb{R}^{N},y\in \mathbb{N}\}$, denoted as $\text{VII}(L,f,\mathcal{D})$ is defined as follows
\begin{equation}
   \begin{aligned}
   \text{VII}(L,f,\mathcal{D})=\mathbb{E}_{(x,y)\sim\mathcal{D}}\frac{|\left<d_x,g_x\right>|}{||d_x||\cdot ||g_x||}\\
   d_x = \text{sign}(x-\bar{x})\\g_x = \text{sign}(\nabla_x L(f(x),y)),
   \end{aligned}
\end{equation}
where $\bar{x}$ denotes the average of $x$ over the whole dataset.
\end{definition}

A higher VII indicates that the model has a more interpretable input-gradient aligned with input over the test set under the certain loss. In the following sections, we will use cross-entropy as the loss function to calculate this metric.
The input-gradients will be scaled to $[0,1]$ for visualization.

\subsection{Effectiveness of Noise Modulation}\label{sec:effectiveness}

\begin{figure*}[t]
   \begin{center}
      \includegraphics[width=0.9\linewidth]{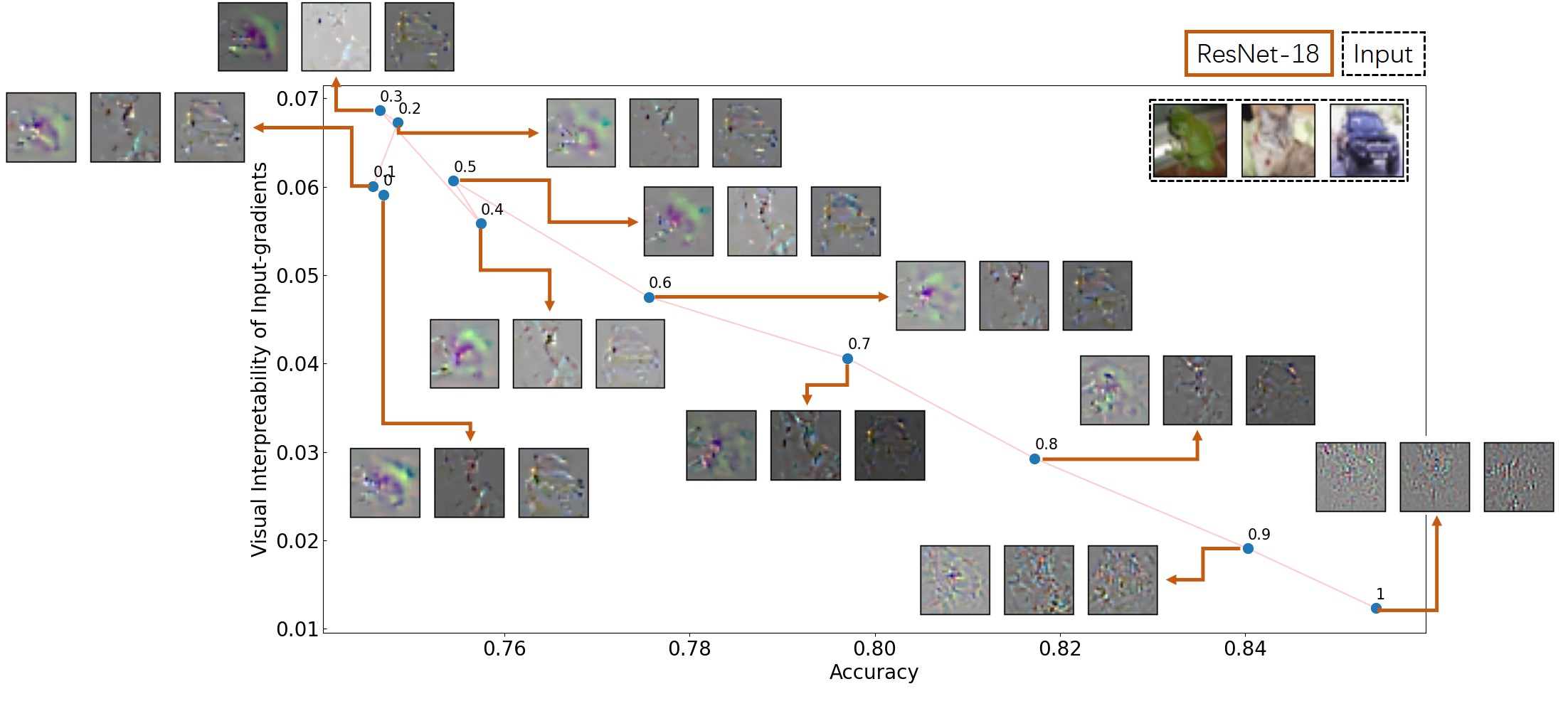}
   \end{center}
   \caption{Trade-off between visual interpretability of input-gradients and accuracy.
   The dotted frame marks the inputs.
   The effects of \emph{noise modulation} plateaus and resonates when $\beta$ is smaller than $0.6$.
   When $\beta=0.8$, the input-gradients become just visually interpretable.}
   \label{fig:vis_tradeoff}
\end{figure*}

\begin{figure*}[t]
   \begin{center}
      \includegraphics[width=0.9\linewidth]{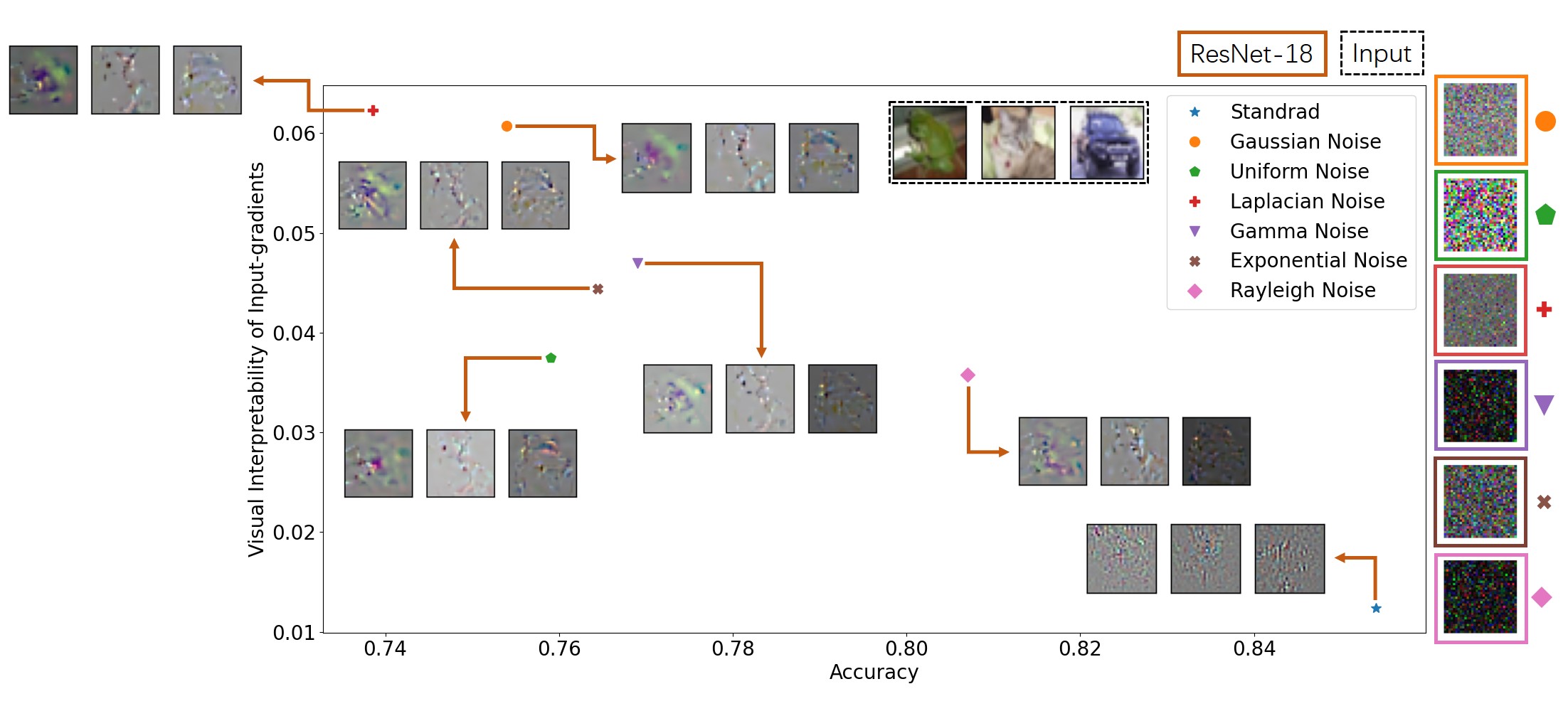}
   \end{center}
   \caption{The influences of noises.The dotted frame marks the inputs.
   The corresponding noises are scaled to $[0,1]$ and visualized on the right with corresponding colors and markers. 
   All of these noises can acquire an interpretable input-gradients.
   Gaussian noise has the best overall performance as expected. 
   Rayleigh noise maintains most of the accuracies with more compromisement of interpretability.
   Laplacian noise achieves the best interpretability, as well as the lowest accuracy.}
   \label{fig:vis_noise}
\end{figure*}

In this section, we will demonstrate the effectiveness of noise modulation over input-gradients.
We will evaluate our method on MNIST~\cite{Lecunmnist} and CIFAR-10~\cite{Alex2009cifar10}.
We will use two simple models, a three layer Multi-Layer Perceptron (MLP) and a six layer Fully Convolutional Neural Networks(FCNN)
on all of these datasets.
Besides, we will also evaluate LeNet~\cite{Lecun1998lenet} on MNIST, VGG~\cite{Karen2015vgg} and ResNet~\cite{Kaiming2016resnet} on CIFAR-10.

Since the focus of this paper is \textbf{NOT} the state-of-art performance of these models, 
for fair comparison, we will use a basic setting without fine-tuning over each models.
All of these models are trained from scratch using a mini-batch size of 64 for 50 epochs.
The optimizer is Adam~\cite{Diederik2015Adam} with a fixed learning rate of 0.001.

All of the inputs are first scaled to $[0,1]$ and prepossessed with normalization, 
no extra data augmentation is used. 
For noise modulation, the ratio of constant component $\beta$ is set to $0.5$ and 
the noise used as carrier is a standard Gaussian noise.
The model with highest validating accuracy during training is saved for comparison.

The experiment results are presented in Figure \ref{fig:vis_comparison}.
We observe that standard models already produce basically interpretable input-gradients for human on MNIST, but 
on CIFAR-10, these standard models produce input-gradients with irregular noises that make no sense for human.
\emph{Noise modulation} significantly increases the interpretability of input-gradients with a few accuracy cost, 
except for MLP on MNIST.
The MLP trained standardly on MNIST already has an input-gradient with most of its non-zero components 
centered around the digits, \emph{noise modulation} focuses its gradients around some critical points on digits 
but leaves a more noisy background that reduces the value of VII metric.

\begin{figure*}[t]
   \begin{center}
      \includegraphics[width=0.9\linewidth]{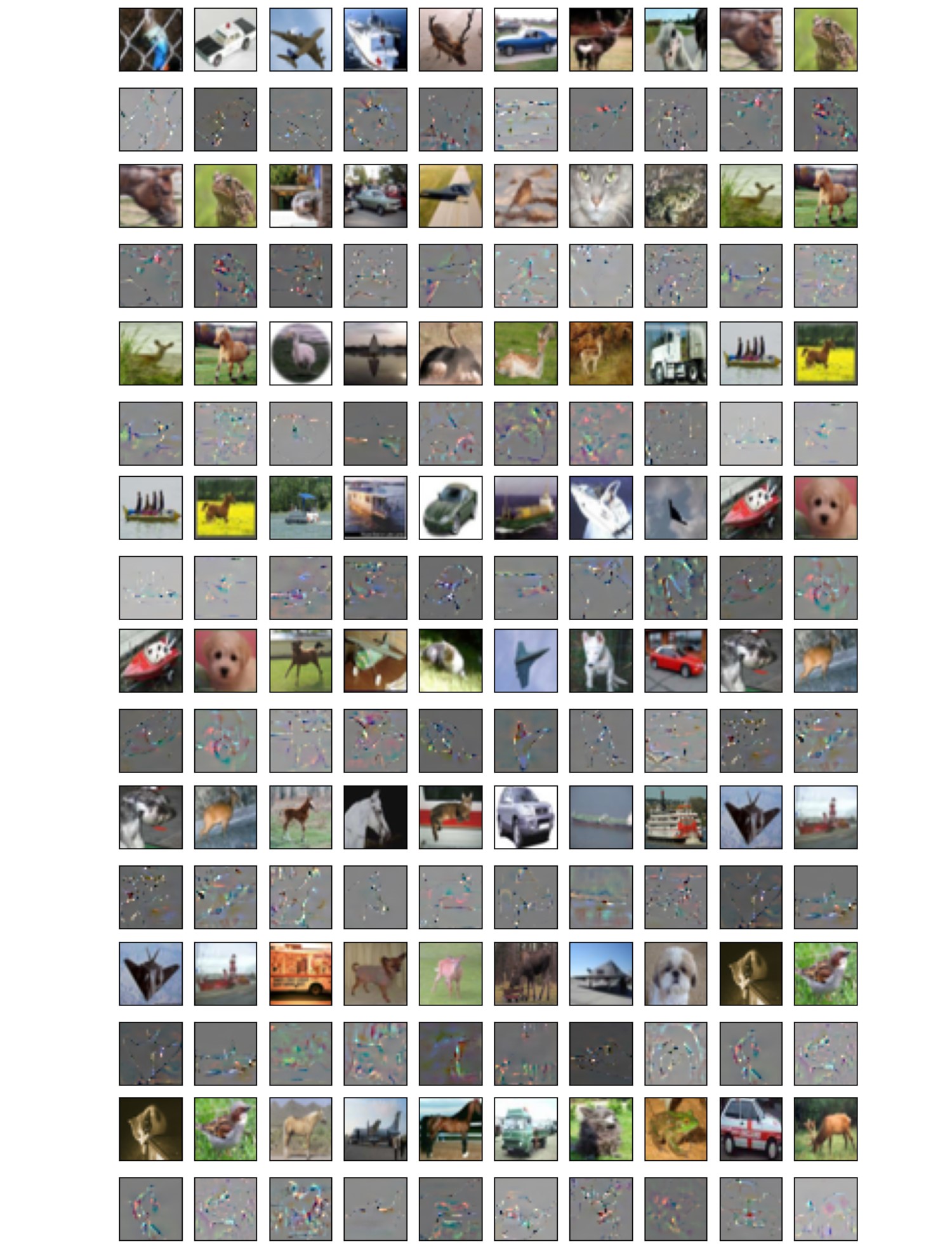}
   \end{center}
   \caption{Visualization of 80 more input-gradients with corresponding inputs for the model trained with \emph{noise modulation} in Section \ref{sec:effectiveness}.}
   \label{fig:vis_more}
\end{figure*}

We also observe that a larger model is more capable of producing interpretable input-gradients, 
but a simple FCNN is also able to produce a reasonable input-gradient for human.
It is clear that \emph{noise modulation} is effective on increasing the interpretability of input-gradients.
More results can be found in Figure \ref{fig:vis_more} at the end of this paper.

\subsection{Trade Interpretability for Accuracy}

While the interpretability increases, \emph{noise modulation} still brings a drop of accuracy.
The design of constant component in Definition \ref{def:noise_modulation} is meant to deal with this problem.

On the same ResNet-18 we use in Section \ref{sec:effectiveness}, we will change the ratio of constant component $\beta$ 
in carrier and check its influences on accuracy and interpretability.
Theoretically, a larger $\beta$ indicates a noisier input-gradient and a higher accuracy as
\emph{noise modulation} becomes standard training when $\beta=1$, 

The experiment results are presented in Figure \ref{fig:vis_tradeoff}.
As expected, a larger $\beta$ yields a higher accuracy and a lower visual interpretability of input-gradients.
From human perception, the input-gradients become roughly interpretable when $\beta$ is smaller than $0.8$.
Reducing $\beta$ makes the interpretability increase until it plateaus and resonates 
once the constant component of carrier is smaller than $0.5$ and the noise dominates carrier.

Given these experiment results,  
we recommend to set the constant component ratio $\beta$ as $0.8$ for a basically interpretable input-gradient 
and $0.5$ to exploit the best interpretability over a compromisement with accuracy.

\subsection{Choice of Noises}

The choice of noises is another hyper-parameter in \emph{noise modulation}.
Intuitively, a Gaussian noise spanning the whole spectrum is the proper choice,
as we intend to have the model learn to filter out the informative features by itself, 

Besides standard Gaussian noise, we will test other five noises on the same ResNet-18 in Section \ref{sec:effectiveness}
with every condition fixed but the type of noise.
These noises include Uniform noise, Laplacian noise, Gamma noise, Exponential noise and Rayleigh noise.
The Uniform noise is sampled from $[0,1]^N$. 
The distribution of Laplacian noise peaks at $0$, and the exponential decay of it is set as $1$.
The shape of the distribution of Gamma noise is set to $1$, as well as its scale.
For Exponential noise and Rayleigh noise, the scale is also set to $1$.

The experiment results are presented in Figure \ref{fig:vis_noise}.
All of these noises yield a much more interpretable input-gradients.
As expected, Gaussian noise has the best overall performance. 
The accuracy loss brought by Rayleigh noise is the least, as well as the 
increasement of interpretability.
Laplacian noise brings the highest increasement of interpretability 
but sacrifies most of the accuracies too.

\section{Related Work}\label{sec:relatedworks}

\textbf{Adversarial Training and Interpretability}
Adversarial training was initially proposed by ~\cite{Ian2015explaining} 
as a defense against adversarial attack, 
but the most popular and effective variation is the PGD adversarial training proposed by ~\cite{Madry2018towards}. 
The discovery that adversarial training results a more interpretable input-gradient was 
first mentioned in ~\cite{Madry2019robustnessodds}.  
~\cite{Tianyuan2019intrepreting} gave an empirical interpretation of the adversarially trained model and further confirmed the increased interpretability brought by adversarial training. 
More theoretical analysis about the mechanism of adversarial training can be found in ~\cite{Zeyuan2020feature}.
~\cite{Christian2019connection} ~\cite{kim2019bridging} discussed the connection between adversarial 
robustness and interpretability.
~\cite{dong2017towards} tried to improve the interpretability of model using adversarial training.
~\cite{Alvin2020Jacobian} ~\cite{Andrew2018inputreg} ~\cite{chan2020thinks} tried 
to utilize this similarity between input-gradients and inputs to increase adversarial robustness.
To our knowledge, we make the first step to exploit this similarity to increase the interpretability of models without adversarial training.

\textbf{Input-gradients and Interpretability} The existence of tiny adversarial perturbations 
found easily through gradient ascent~\cite{Christian2014intriguing} indicates that input-gradients of model 
contain information critical to the predictions of model, 
but the vanilla input-gradients of a standard model generally gives no interpretable information for human~\cite{Mukund2017axiomatic}. 
Many interpreting methods try to process vanilla gradients with some extra operations to give an interpretable heatmap for human, 
including multiplying the gradient with input(Gradient*Input method~\cite{Avanti2016gradinput}),  
integrating the gradients (Integrated Gradients ~\cite{Mukund2017axiomatic}),  
averaging gradients of perturbed inputs (SmoothGrad~\cite{Daniel2017smoothgrad}), 
combining gradients with features (GradCAM~\cite{Ramprasaath2017gradcam}) etc.
All of these heatmaps, except for GradCAM and vanilla gradients, have been shown by ~\cite{Julius2018sanitychecks} that 
the results are independent from both data, model and model parameters.
However, the input-gradient has a higher resolution compared with GradCAM, once made interpretable,
it gives a more precise explanation for the model's decision.

\section{Conclusion}\label{sec:conclusion}

In this paper, we connect adversarial training with \emph{amplitude modulation}
through an approximation of adversarial perturbations. 
It offers a new perspective to understand the effects of adversarial training on input-gradients.
Based on a digital analogy of \emph{amplitude modulation},
we propose \emph{noise modulation} as an efficient and model-agnostic alternative to 
train a model that interprets itself with clear and human-aligned input-gradients.
We confirm the effectiveness of \emph{noise modulation} on input-gradients, 
draw a trade-off between interpretability and accuracy and analyze the effects of different choices of noises. 
We believe this work will serve as a baseline towards efficient training strategies that grants model with interpretability without hurting its ability.

{\small
\bibliographystyle{ieee_fullname}
\bibliography{nmbib}
}

\end{document}